\documentclass{article}
\usepackage{fancyhdr}
\usepackage{spconf,amsmath,graphicx}
\usepackage{spconf,amsmath,graphicx}
\usepackage{graphicx}
\usepackage{amsmath}
\usepackage{amssymb}
\usepackage{booktabs}

\usepackage{soul}
\usepackage{changes}
\definechangesauthor[name={Percusse}, color=orange]{per}

\usepackage{soul}
\usepackage{changes}

\fancypagestyle{firstpage}{
  \fancyhf{}
  \fancyhead[L]{\textit{Proceedings of IC-NIDC 2023}}
  
}

\title{Low Latency Spiking Neural Network for Object Detection}
\name{Nemin Qiu, Chuang Zhu$^{\ast}$}
\address{School of Artificial Intelligence, Beijing University of Posts and Telecommunications.\\
czhu@bupt.edu.cn}
\begin{document}
%
\maketitle
\thispagestyle{firstpage}

\begin{abstract}

Spiking Neural Networks (SNNs), as a third-generation neural network, are well-suited for edge AI applications due to their binary spike nature. However, when it comes to complex tasks like object detection, SNNs often require a substantial number of time steps to achieve high performance. This limitation significantly hampers the widespread adoption of SNNs in latency-sensitive edge devices. In this paper, our focus is on generating highly accurate and low-latency SNNs specifically for object detection. Firstly, we systematically derive the conversion between SNNs and ANNs and analyze how to improve the consistency between them: improving the spike firing rate and reducing the quantization error. Then we propose a structural replacement, quantization of ANN activation and residual fix to allevicate the disparity. We evaluate our method on challenging dataset MS COCO, PASCAL VOC and our spike dataset. The experimental results show that the proposed method achieves higher accuracy and lower latency compared to previous work Spiking-YOLO. The advantages of SNNs processing of spike signals are also demonstrated.

\end{abstract}
\section{Introduction}
\label{sec:intro}

Artificial neural networks have achieved great
success in computer vision \cite{redmon2016you}, natural
language processing, and other domains. Despite these achievements, there still exists a fundamental difference between the operational mechanisms of artificial neural networks and human neural activity. Consequently, some researchers have begun studying neural networks that emulate the neural activity of the human brain. Spiking neural networks (SNNs) are considered as the third generation of neural network models, utilizing simplified yet biologically realistic neuron models for computation. SNNs differ from traditional artificial neural networks, such as convolutional neural networks (CNNs), in that they transmit activation data between layers as sequences of binary spikes, following specific firing rules. SNNs significantly reduce computational resource requirements and effectively avoids excessive resource consumption \cite{kim2020spiking}. As SNNs have demonstrated successful applications in edge AI \cite{kim2020spiking}, research in this field is gaining increased attention from researchers.

In general, there are two mainstream methodologies for
developing deep supervised SNNs up to date: direct
training for SNNs and converting ANNs into SNNs.
However, directly trained SNNs generally do not achieve better performance on relatively complex vision scenes and tasks \cite{kim2020spiking}. For directly trained SNNs, on the one hand,the back-propagation algorithm could
not be applied to SNNs as spiking activation functions are inherently non-differentiable. It is hard to find a way to update the SNN neural network weights well. This leads to difficulties in achieving satisfactory performance of SNN on tasks with complex scenarios. On the other hand, directly trained SNNs usually use complex neuron models without specified optimization for storage and operation with binary events. This leads to less practicality\cite{kim2020spiking}.

For converted SNNs, as they are transferred from a certain pre-trained ANN model, it is possible to make the SNN achieve a performance close to that of the ANN. in order to attain sufficient representation precision, a considerable number of time steps are usually required for a nearly lossless conversion, which is known as the accuracy-delay tradeoff. This tradeoff significantly restricts the practical application of SNNs. The consumption of a large number of time steps can result in a significant delay in SNN inference, which is detrimental for certain real-time tasks, such as the object detection task emphasized in this paper. Recent works \cite{li2021free} \cite{ding2021optimal} propose methods to alleviate this problem by exploiting the quantization and clipping properties of aggregation representations. However, these works primarily focus on the image classification task and overlook the impact of residual voltage and neuron firing rate on error propagation. There still exists a noticeable performance gap between ANNs and SNNs when it comes to low inference latency, and the underlying cause for this degradation remains unclear.

In this work, we identify that the conversion error under low time steps mainly arises from low spike firing rate, quantization error, and the misrepresentation of residual membrane potential. These factors accurately characterize the information loss between the input and output of spiking neurons with asynchronous spike firing.
Inspired by these findings, we propose methods to address these issues, namely the low spike firing rate layer replacement, quantization activation, and residual fix methods. By implementing these techniques, we generate an SNN for object detection that achieves remarkable performance with an extremely low inference delay. The main contributions of this work can be summarized as follows:
\begin{itemize}
\item We describe the specific conversion process of ANNs to SNNs and model the errors introduced during the ANN-SNN conversion. Then we propose methods to reduce these errors.

\item  We propose a scheme for layer replacement in the low spike firing rate layer and quantized ANNs activation for the adaptation conversion. In the first phase, some SNNs-unfriendly layers were replaced in the ANNs before conversion. Quant-ReLU functions are applied to finetune ANNs. In the second phase, using residual fix mechanisms in IF neurons.

\item We verify the effectiveness and efficiency of the proposed methods on the MS COCO, PASCAL VOC and spike dataset. Experimental results show significant improvements in accuracy-latency tradeoffs compared to previous works.

\end{itemize}

\section{Related work}
\label{sec:format}

Existing SNNs are generally divided into two fields
to study, directly trained SNNs and converted SNNs \cite{sengupta2019going}. For directly trained SNNs, unsupervised and supervised learning
are both attractive research topics. On the one hand unsupervised learning, the mainstream learning method is the
spike timing dependent plasticity rule (STDP). STDP uses synaptic plasticity and spike activity to learn features of input data, which is biologically
plausible. On the other hand, the supervised SNNs can
achieve much better performance given a large number of
labeled training data. There are some successful attempts
that introduce BP into SNN models, such as STBP, SLAYER, BPSTDP, which achieve good performance on some simple cognitive tasks.

The conversion of ANN-SNN is in burgeoning research. Cao et al. \cite{cao2015spiking} proposed a ANN-SNN conversion method that neglected bias and max-pooling. In the next work, Diehl et al \cite{diehl2015fast} proposed data-based normalization to improve the performance in deep SNNs. Rueckauer et al \cite{rueckauer2017conversion} presented an implementation method of batch normalization and spike max-pooling. Sengupta et al \cite{sengupta2019going} expanded conversion methods to VGG and residual architectures. Nonetheless, most previous works have been limited to the image classification task \cite{rueckauer2017conversion}. Kim et al \cite{kim2020spiking} have presented Spiking-YOLO, the first SNN model that successfully performs objectdetection by achieving comparable results to those of the original DNNs on non-trivial datasets, PASCAL VOC and MS COCO. Ding et al \cite{ding2021optimal} presented Rate Norm Layer to replace the ReLU function, and obtain the scale through a gradient-based algorithm. Conversion approaches have revealed their potential of achieving ANN level performance in various tasks \cite{kim2020spiking}. However, taking advantage of the artificial neural network’s success, network conversion outperforms other methods without auxiliary computation power involved. converted SNN model suffers from efficiency problems. SNNs converted require massive timesteps to reach competitive performance \cite{kim2020spiking}. All of them are complicated procedures vulnerable to high inference latency \cite{kim2020spiking}. Converted SNNs still suffer from increased energy consumption, long inference time and high time delays \cite{kim2020spiking}. Building on these previous efforts, We put toward to minimize the ANN-to-SNN conversion error in complex visual scene tasks, using ultra-low latency when achieving high precision SNNs.

\section{Preliminaries}
In this section, we introduce the activation propagation rule of ANNs and the working principle of the spiking neurons.

\textbf{ANN:}
Let $x$ denote the activation. The relationship between the activation of the two adjacent layers is as follows:
\begin{equation}
  x\sb{k}^{l}=f(\sum_{j}(w\sp{l}\sb{k,j}\cdot\,x\sb{j}^{l-1})+b\sb{k}\sp{l}),
  \label{eq:ANN_activations}
\end{equation}
where $w\sp{l}\sb{k,j}$ represents weights, $b\sb{k}\sp{l}$ represents biases, and the activation $x\sb{k}^{l}$ represents the activation of neuron $k$ in layer $l$. $f(.)$ is a type of activation function. 

\textbf{SNN:}
  Let us focus on the IF neuron model. Let $U_{k}^{l}(t)$ denote a transient membrane potential increment of spiking neuron $k$ in layer $l$:

\begin{equation}
  U_{k}^{l}(t)=\sum_{j}(w_{k,j}^{l}\cdot \Theta_{t,j}^{l-1})+b\sb{k}\sp{l},
  \label{eq:transientmembrane}
\end{equation}
where $\Theta_{t,k}^{l}$ denotes a step function indicating the occurrence of a spike at time $t$:
\begin{equation}
  \Theta_{t,k}^{l}=\Theta(V_{k}^{l}(t-1)+U_{k}^{l}(t)-V_{k,th}),
  \label{eq:Theta_in_SNN}
\end{equation}
\begin{equation}
  ~\mathrm{with}~\Theta(x)=\left\{\begin{array}{l l}{{1,}}&{{x\ge{0}}}\\ {{0,}}&{{\mathrm{otherwise}}}\end{array},\right.
  \label{eq:Theta_whether_fire}
\end{equation}

The spiking neuron integrates inputs $U_{k}^{l}(t)$ until the membrane potential $V_{k}^{l}(t-1)$ exceeds a threshold $V_{k,th}$ and a spike is generated. After a spike is fired at time $t$, the membrane potential is reset. The formula for resetting $V_{k}^{l}(t)$ is as follows:
\begin{equation}
  V_{k}^{l}(t)=V_{k}^{l}(t-1)+U_{k}^{l}(t)-V_{k,th}\Theta_{t,k}^{l}.
  \label{eq:subtraction mechanism}
\end{equation}

\section{Method}
\begin{figure}
  \centering
  \includegraphics[width=0.5\textwidth]{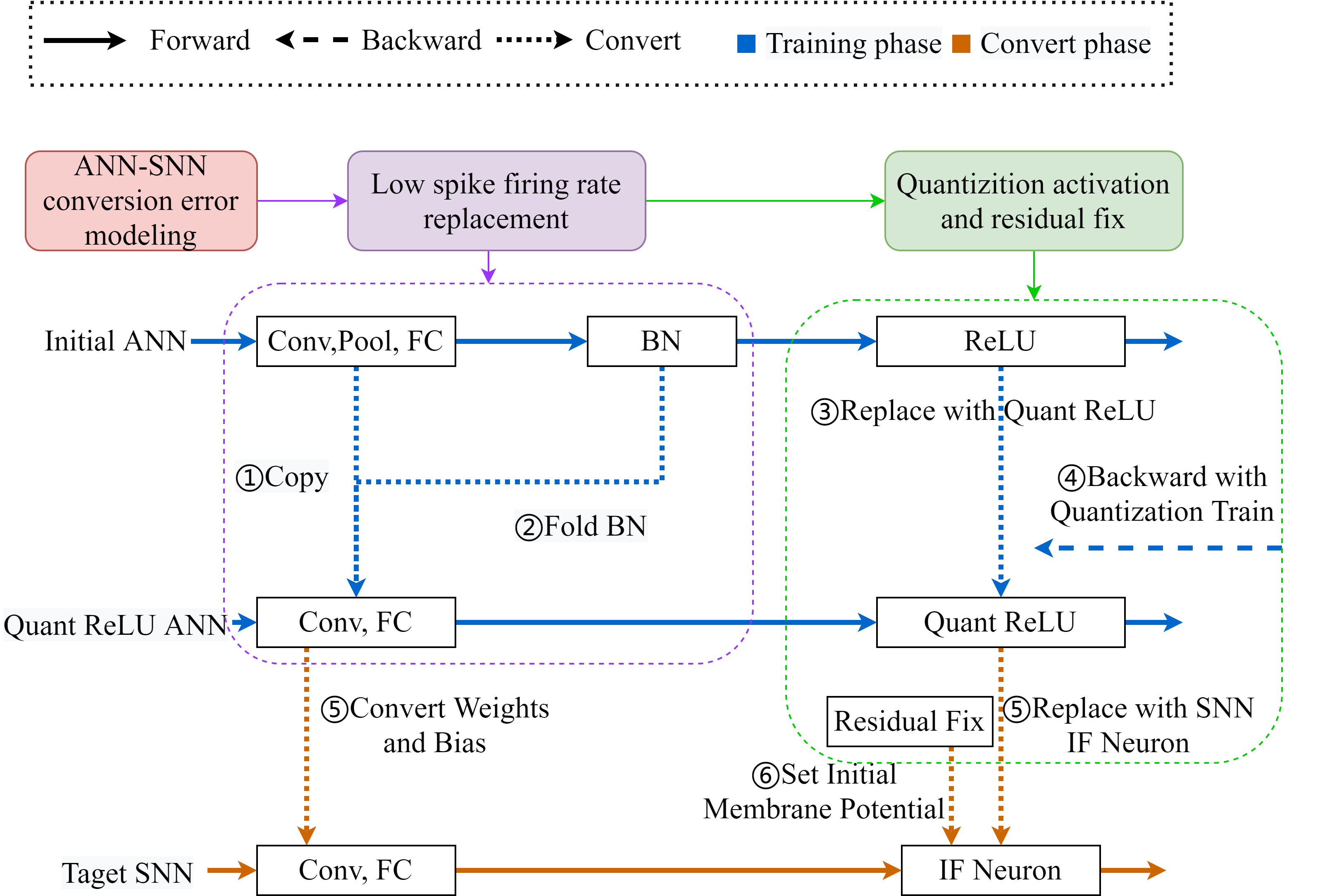}
  \caption{ Overall flow of training phase and converting phase.}
  \label{ANNTOSNN}
  \vskip -0.1 in
\end{figure}

In order to solve the problem of high latency of SNN for object detection task, this paper proposes a method to generate low latency object detection SNN network.
Figure \ref{ANNTOSNN} illustrates the overall flow of the generation. The 'unfriendly' modules in the ANN network are first replaced,then the BN layer is fused into the convolution or linear layer. After completing the restructuring of the network, we use Quant-ReLU instead of ReLU, and use quantization training to update the parameters. After completing training, the weights are subsequently converted using the conversion formula \cite{rueckauer2017conversion}, the original activation function is replaced using IF neurons, and finally residual fix is added to further reduce the conversion error. This section describes these improvements that we have proposed through our ANN-SNN conversion error modeling.

\label{sec:format}

\subsection{ANN-SNN conversion error modeling}
\label{sec:3.1}

In this section, we will introduce the proof of ANN-SNN conversion\cite{rueckauer2017conversion}.And from it, we analyze some improvements that will help reduce latency and improve performance after conversion.

To simplify the description, we assume that the interval of time step $d t = 1$, and inferring an image takes $T$ time steps. The firing rate of each SNN
neuron as $r_{k}^{l}(T)=N_{k}^{l}(T)/T$, where $N_{k}^{l}(T)=\sum_{t=1}^{T}\,\Theta_{t,k}^{l}$ is the number of spikes generated. From the above definition, it is clear that the firing rate of IF neurons $r_{k}^{l}(T)\,\in\,[0,1]$. Moreover, it is clear that the firing rate is discrete, and has a resolution of 1/T.

Assume that the initial membrane potential is zero $V_{k}^{l}(0)=0$. After accumulating T time steps, the membrane  potential at any time point $T$ is given as $V_{k}^{l}(T)= \sum_{t=1}^{T}\,U_{k}^{l}(t)-V_{k,th}\cdot N_{k}^{l}(T)$. From this we can deduce $N_{k}^{l}(T)=\lfloor {\frac{\sum_{t=1}^{T}\,U_{k}^{l}(t)-V_{k}^{l}(T)}{V_{k,th}}}\rfloor$ 
, and then the firing rate  $r_{k}^{l}(T)$ is:
\begin{equation}
  r_{k}^{l}(T)=\frac{N_{k}^{l}(T)}{T}=\frac{\lfloor {(\frac{\sum_{t=1}^{T}\,U_{k}^{l}(t)}{{V_{k,th} \cdot T}}-{\frac{V_{k}^{l}(T)}{V_{k,th} \cdot T}}) \cdot T}\rfloor}{T} .
  \label{eq:e5}
\end{equation}

\begin{equation}
  r_{k}^{l}(T)=\frac{N_{k}^{l}(T)}{T}=\frac{\sum_{t=1}^{T}\,U_{k}^{l}(t)}{{V_{k,th} \cdot T}}-{\frac{V_{k}^{l}(T)}{V_{k,th} \cdot T}}.
  \label{eq:e11}
\end{equation}

  Assume that the threshold $V_{k,th}=1$. With this subtraction mechanism, let the Eq. (\ref{eq:e5}) change to:
\begin{equation}
  r_{k}^{l}(T)=\frac{\lfloor {(\sum_{j}(w_{k,j}^{l}\cdot {\frac{\sum_{t=1}^{T}\,\Theta_{t,j}^{l-1}}{T}})+b\sb{k}\sp{l}-{\frac{V_{k}^{l}(T)}{ T}}) \cdot T}\rfloor}{T} ,
  \label{eq:r2}
\end{equation}

\begin{equation}
  r_{k}^{l}(T)= \frac{\lfloor {(\sum_{j}(w_{k,j}^{l}\cdot {r_{j}^{l-1}(T)})+b\sb{k}\sp{l}-{\frac{V_{k}^{l}(T)}{T}}) \cdot T}\rfloor}{T}.
  \label{eq:r4}
\end{equation}
Here we define an approximation of the firing rate $\widetilde{r}_{k}^{l}(T)$:

\begin{equation}
  \widetilde{r}_{k}^{l}(T)= \sum_{j}(w_{k,j}^{l}\cdot {r_{j}^{l-1}(T)})+b\sb{k}\sp{l}-{\frac{V_{k}^{l}(T)}{T}}.
  \label{eq:r_appp}
\end{equation}
The relationship between the approximation and the true value of  firing rate is:

\begin{equation}
  r_{k}^{l}(T)= \frac{\lfloor {(\widetilde{r}_{k}^{l}(T)) \cdot T}\rfloor}{T}.
  \label{eq:r_comp_rapp}
\end{equation}

\begin{figure}
  \centering
  \includegraphics[width=0.44\textwidth]{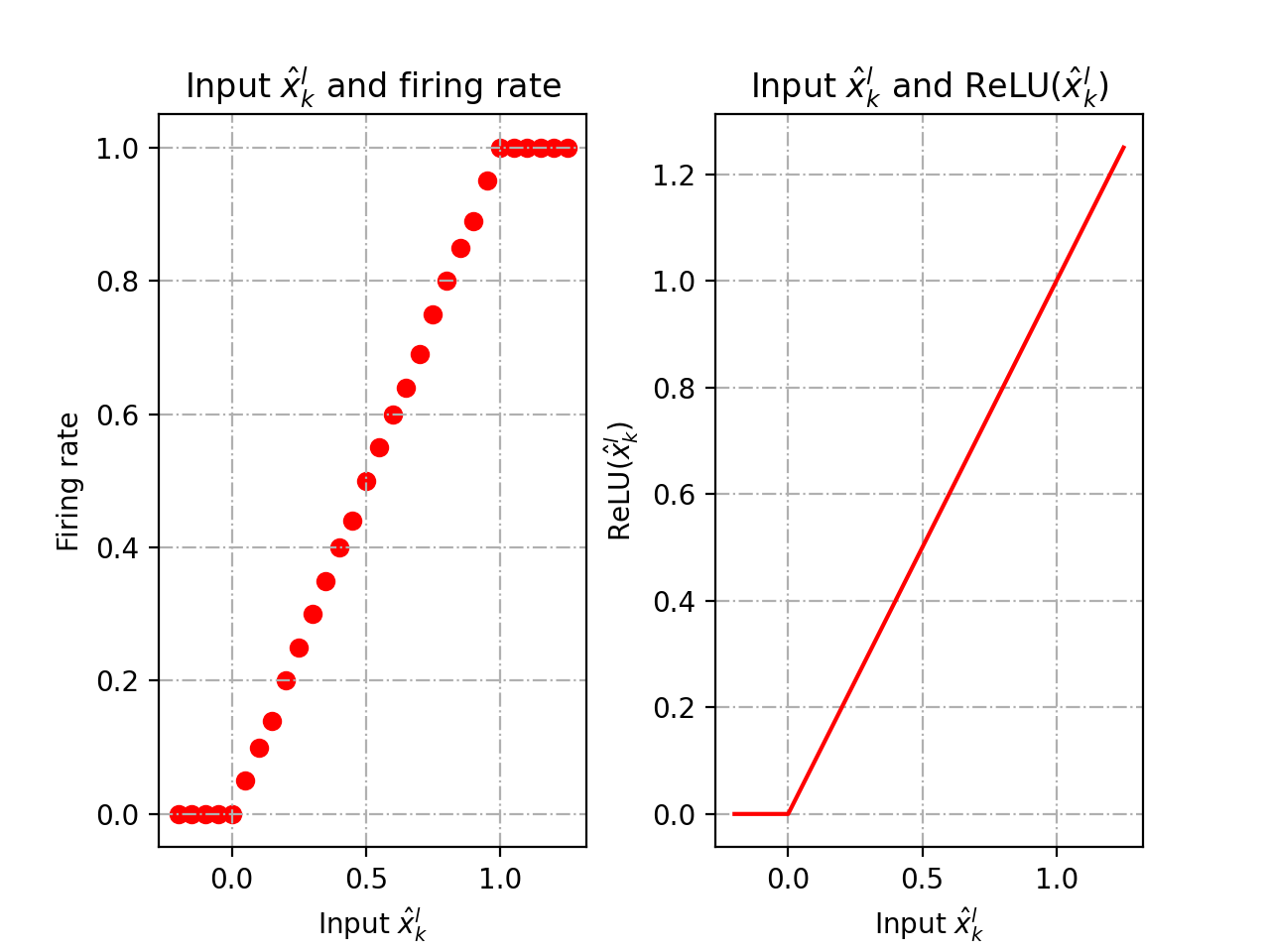}
  \caption{Firing rate of IF neuron and ReLU.}
  \label{firing_rate}
  \vskip -0.1 in
\end{figure}

\textbf{ANN to SNN}: 
The similarity of IF neurons and ReLU activation functions is an important basis on which ANNs can be converted to SNNs. The principle of ANN-SNN conversion is that the firing rates of spiking neuron $r_{k}^{l}(T)$ should correlate with the original ANN activations $x_{k}^{l}$ such that $r_{k}^{l}(T){\rightarrow} x_{k}^{l}$. Setting $V_{k}^{l}(0)=0$ , $V_{k,th}=1$. Figure \ref{firing_rate} shows the corresponding relationship between the output of ReLU activation function in ANN and the output of firing rate in SNN with the input $\hat{x}_{k}^{l}=\sum_{j}(w\sp{l}\sb{k,j}\cdot\,x\sb{j}^{l-1})+b\sb{k}\sp{l}$ and $ r_{k}^{l}(T)
= \sum_{j}(w_{k,j}^{l}\cdot {r_{j}^{l-1}(T)})+b\sb{k}\sp{l}$.  

\subsection{Low spike firing rate layer replacement method }
\label{sec:3.2}

After the derivation in Section \ref{sec:3.1}, we can know that the purpose of the conversion is to make $ r_{k}^{l}(T) = \sum_{j}(w_{k,j}^{l}\cdot {r_{j}^{l-1}(T)})+b\sb{k}\sp{l}$. Let us define that the following $\frac{V_{k}^{l}(T)}{ T}=Re$ is a remainder. When the time steps T is very large, the remainder $Re=\frac{V_{k}^{l}(T)}{ T}\approx0$ and $ r_{k}^{l}(T)\approx\widetilde{r}_{k}^{l}(T)$. The other case is that the value of the membrane potential $V_{k}^{l}(T)$ is exactly 0 after each T time step. In this case, $Re=\frac{V_{k}^{l}(T)}{T}=0$. In addition, the remainder relative error is reduced if the first half of Eq. (\ref{eq:r_appp}) has a larger value. It can be understood as $r_{k}^{l}(T)$ has a larger spike firing rate. After satisfying the above conditions we can conclude that:
\begin{equation}
  r_{k}^{l}(T)\approx\widetilde{r}_{k}^{l}(T)\approx\sum_{j}(w_{k,j}^{l}\cdot {r_{j}^{l-1}(T)})+b\sb{k}\sp{l}.
  \label{eq:r3}
\end{equation}

After the above analysis, it is known that the low spiking firing rate introduces a large error in the ANN-SNN conversion method. The previous works did not bother about the impact that the modules in the network structure bring to the ANN-SNN conversion, we counted the spike firing rate of each layer, and we found that the spike firing rate decreased very seriously in the maximum pooling layer. However, the down-sampling operation inevitably causes the loss of spike information in the convolutional feature map, which affects the detection accuracy. Down-sampling of spike information requires a more accurate information integration process. Considering this, we modified the original network model structure by replacing the network structure of the original model using downsampling convolution and transposed convolution. After the replacement and conversion, we found some improvement in the spike firing rate of neurons in this layer, the time step required for the original model to achieve the same accuracy is shorter.

\subsection{Quantizition activation and residual fix methods }
\label{sec:3.3}

\begin{figure}
  \centering
  \includegraphics[width=0.5\textwidth]{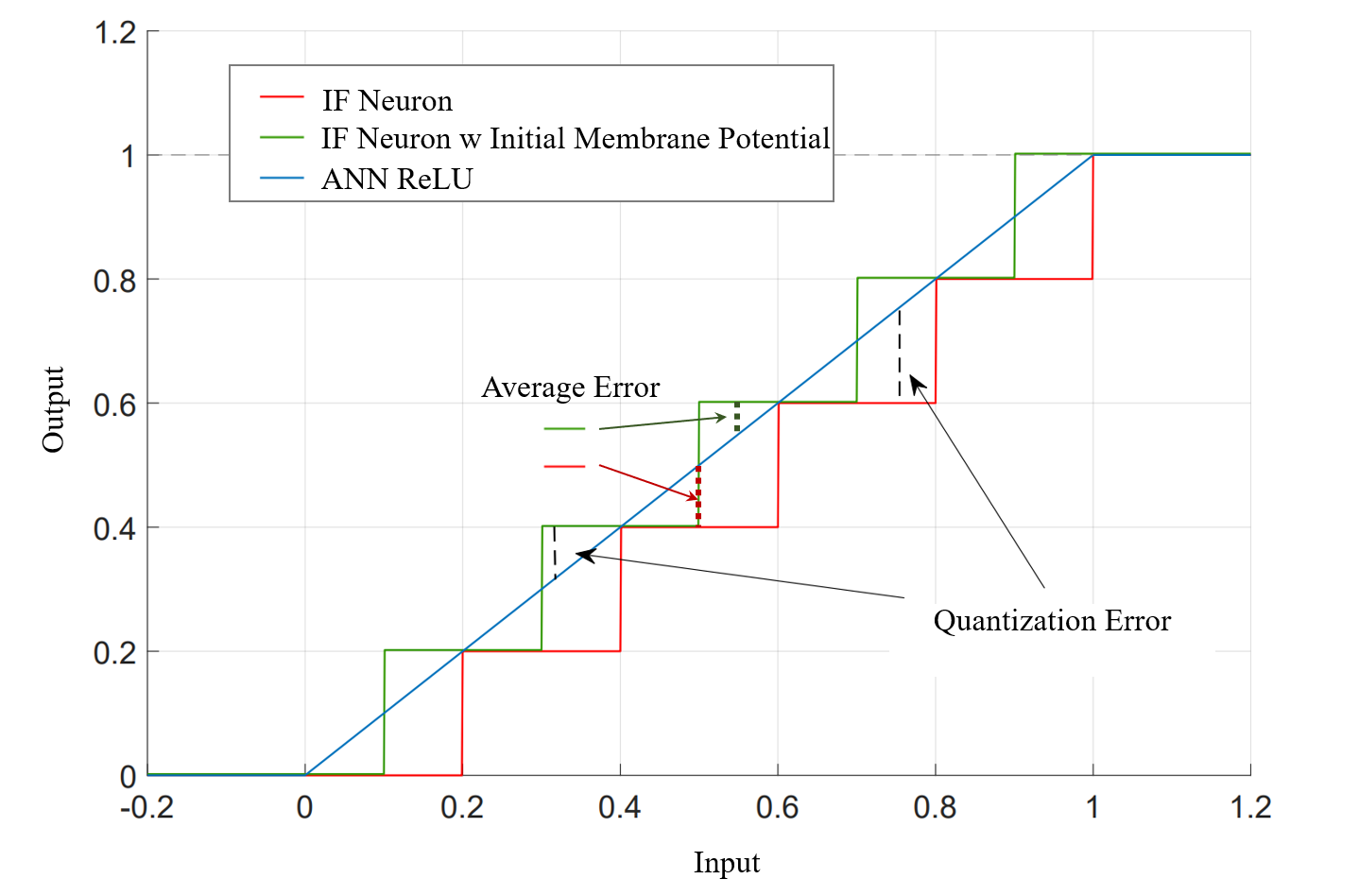}
  \caption{The firing rate of the IF neuron and ANN ReLU.}
  \label{quant-relu}
  \vskip -0.1 in
\end{figure}

In addition to the analysis of spiking firing rate on the network structure. According to equations Eq. (\ref{eq:r_appp}) and Eq. (\ref{eq:r_comp_rapp}), we also analyzed that there is a quantization error due to the gap between the two activation patterns during the conversion, Figure \ref{quant-relu} shows the activation resolution of the spiking neural network is 1/T (T is the time step). Obviously if reducing the gap between the expression of the two neurons helps to reduce the conversion error. For the SNN, we added a residual recovery setting. To address the residual error, we set a specific initial value for the membrane potential when configuring it. This helps to reduce the difference between the activation forms of IF neurons and ReLU neurons in the ANN network. For the ANN,  We set the activation value of the activation function ReLU of ANN to be discrete and the resolution is also 1/T. so we proposed a strategy of activation substitution in the training phase and setting the initial membrane potential to reduce the error between the two. 

Specifically, our scheme is using Quant-ReLU activation (quantization clipping function) instead of ReLU activation in the pre-transformation ANN training. learning and continuously reducing the quantization error through powerful ANN training methods, making the activation form of the SNN simulated in the training phase. Quant-ReLU activation is more similar to the activation form of IF neurons of the SNN. For residual fix, we set the initial membrane potential $V_{k}^{l}(0)=0.5$. This setting makes the output firing rate of IF neurons and the input firing rate as shown in the green line in Figure \ref{quant-relu}, which can greatly reduce the quantization error. In terms of theoretical calculations, the average error ratio of the two methods (red and green lines) is 2 to 1. Similarly, the activation form of Quant-ReLU is also set to this form, which facilitates the ANN training to get the best performance.

Finally, we need to convert the weights of the ANN according to the weight conversion formula (due to the characteristics of spiking neurons, the spiking frequency cannot be higher than one) \cite{rueckauer2017conversion}. After the weight conversion  as mentioned in main paper, Eq. (\ref{eq:ANN_activations}) changes to the following form:
\begin{equation}
  x\sb{k}^{l}=f(\sum_{j}(\hat{w}\sp{l}\sb{k,j}\cdot\,x\sb{j}^{l-1})+\hat{b}\sb{k}\sp{l}),
  \label{eq:ANN_activations2}
\end{equation}
After weight conversion, the range of $x\sb{k}^{l}$ is [0,1]. According to the definition of ReLU activation function, Eq. (\ref{eq:ANN_activations2}) can then be changed to the following form:
\begin{equation}
  x\sb{k}^{l}=\sum_{j}(\hat{w}\sp{l}\sb{k,j}\cdot\,x\sb{j}^{l-1})+\hat{b}\sb{k}\sp{l},
  \label{eq:ANN_activations3}
\end{equation}
After applying the weights to the SNN ($\hat{w}_{k,j}^{l}{\rightarrow}w_{k,j}^{l}$ and $\hat{b}\sb{k}\sp{l} {\rightarrow} b\sb{k}\sp{l}$), Eq. (\ref{eq:r3}) can then be changed to the following form:
\begin{equation}
  r_{k}^{l}(T)\approx \sum_{j}(\hat{w}_{k,j}^{l}\cdot {r_{j}^{l-1}(T)})+\hat{b}\sb{k}\sp{l}.
  \label{eq:ANN_activations3}
\end{equation}
It can be seen that between two adjacent layers, SNN and ANN pass features with the almost same formula.

\section{Experiment and evaluation}
\label{sec:maintitle}

In this section, all experiments are performed on NVIDIA Tesla V100 32G GPUs and based on the Pytorch framework. For the object detection task, we compare our low-latency SNN detection network with the previous work Spiking-YOLO \cite{kim2020spiking}. Our experiment is tested on MS COCO and PASCAL VOC dataset. In addition we tested on our spike dataset. Our spike dataset is mainly for object detection of people and vehicles in some traffic scenarios. Our spike dataset involved in this work are captured using spiking cameras \cite{zheng2021high} or by encoding the video with spike encoder \cite{kim2020spiking}. The detection results of our experiments are evaluated using mAP50($\%$). 

\begin{table}
\centering
\begin{tabular}{@{}cccc@{}}
\toprule
Method       & Neural Model & Time Steps   & mAP           \\ \midrule
Spiking-YOLO \cite{kim2020spiking} & IBT          & 3500         & 25.66         \\ \midrule
STDP-Spiking \cite{chakraborty2021fully} & IF          & 300         & 26.8         \\ \midrule
Ours         & IF w IMP     & 100          & 14.8          \\ \midrule
Ours         & IF w IMP     & \textbf{150} & 26.3          \\ \midrule
Ours         & IF w IMP     & 200          & 31.5          \\ \midrule
Ours         & IF w IMP     & 250          & 34.2          \\ \midrule
Ours         & IF w/o IMP         & 300          & 35.7          \\ \midrule
Ours         & IF w IMP     & 300          & \textbf{36.2} \\ \bottomrule
\end{tabular}
\caption{Comparison of the object detection performance between our method (Ours) and Spiking-YOLO \cite{kim2020spiking} (using IBT neuron \cite{kim2020spiking}) on PASCAL VOC. (w IMP or w/o IMP means the IF neuron of SNN with or without residual fix.)}
  \label{tab:cocomap}
\end{table}

Table \ref{tab:cocomap} illustrates the significant time step savings of our method over Spiking-YOLO \cite{kim2020spiking} on the dataset MS COCO, we only need 150 time steps to achieve comparable performance, and our method achieves a huge improvement (+10.54) in accuracy at 300 time steps. When comparing our method to STDP-Spiking \cite{chakraborty2021fully}, which utilizes direct training with a similar network structure, we have consistently demonstrated superior performance. Even when considering a time frame of 300 steps, our approach outperforms this method as well. Table \ref{tab:VOCmap} illustrates the significant time step savings of our method over Spiking-YOLO \cite{kim2020spiking} on the dataset PASCAL VOC, we only need 150 time steps to achieve comparable performance, and our method achieves a considerable improvement (+2.37) in accuracy at 300 time steps. By using our method, SNN with better performance can be obtained. To summarize Table \ref{tab:cocomap} and Table \ref{tab:VOCmap}, the inclusion of residual fix in our approach yields a slight yet notable performance gain. 

Experiments show that on MS COCO and PASCAL VOC datasets, SNNs generated using our method can significantly reduce the time step required for inference, with our method requiring only 1/23 of the time step of spiking-yolo to achieve the same accuracy, and our method achieves good performance at 300 time steps. This highlights the effectiveness of our method in pushing the boundaries of SNN capabilities and achieving superior results in neural network applications.

\begin{table}
\centering
\begin{tabular}{@{}cccc@{}}
\toprule
Method       & Neural Model & Time Steps   & mAP           \\ \midrule
Spiking-YOLO \cite{kim2020spiking} & IBT          & 3500         & 51.83        \\ \midrule
Ours         & IF w IMP     & 100          & 48.7          \\ \midrule
Ours         & IF w IMP     & \textbf{150} & 52.0         \\ \midrule
Ours         & IF w IMP     & 200          & 53.1          \\ \midrule
Ours         & IF w IMP     & 250          & 53.8         \\ \midrule
Ours         & IF w/o IMP         & 300          & 53.7          \\ \midrule
Ours         & IF w IMP     & 300          & \textbf{54.2} \\ \bottomrule
\end{tabular}
\caption{Comparison of the object detection performance between our method (Ours) and Spiking-YOLO \cite{kim2020spiking} (using IBT neuron \cite{kim2020spiking}) on MS COCO}
  \label{tab:VOCmap}
\end{table}

\begin{table}[]
\centering
\begin{tabular}{cccc}
\hline
Method & Network         & Time Steps & mAP  \\ \hline
ANN    & Tiny-yolov3     & -          & 51.6 \\ \hline
Ours   & SNN-Tiny-yolov3 & 64         & 45.0 \\ \hline
Ours   & SNN-Tiny-yolov3 & 128        & 50.9 \\ \hline
Ours   & SNN-Tiny-yolov3 & 256        & \textbf{52.4} \\ \hline
\end{tabular}
\caption{Comparison of the object detection performance between our method (Ours) and ANN \cite{redmon2016you} on our dataset. (The input of ANN is the spike signal reconstructed into gray image, and the input of SNN is the spike signal.)}
  \label{tab:ourdatasetmap}
  \vskip -0.1 in
\end{table}

Table  \ref{tab:ourdatasetmap} shows the improvement of SNN compared to ANN for the same network structure on the spike dataset we produced. This dataset is composed of gray images, along with the corresponding spike data and annotation information. This demonstrates that SNNs may have an inherent advantage when processing spike signals.

\begin{figure}[h]
  \centering
  \includegraphics[width=0.48\textwidth]{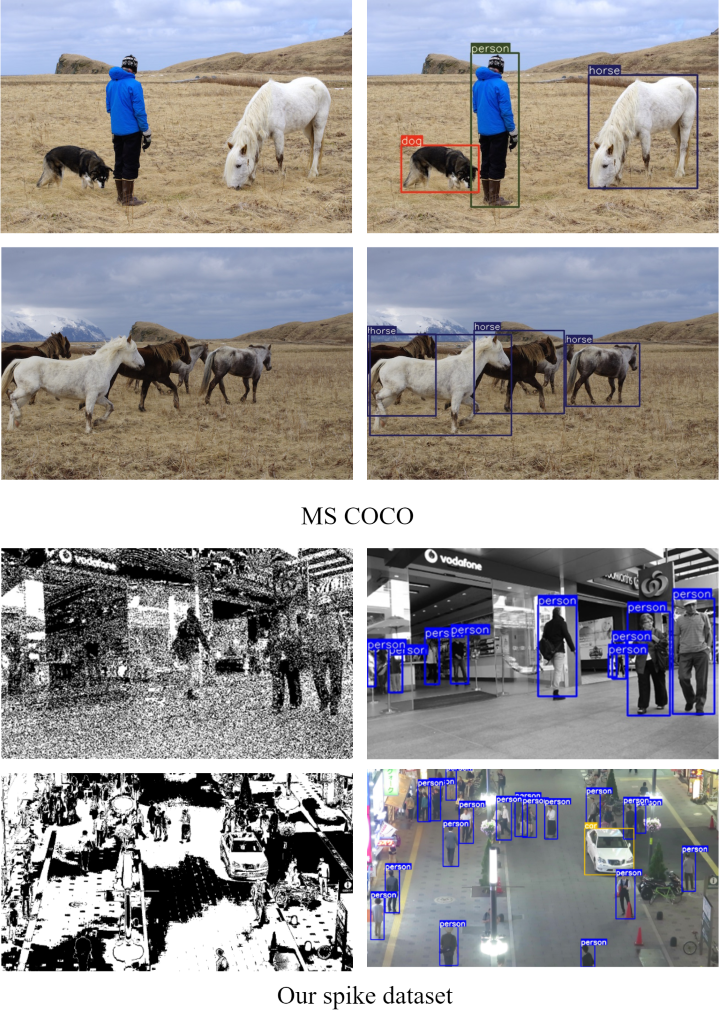}
  \caption{Visualization of object detection results using our method on MSCOCO and our spike dataset. The visualization experiment is set up with 300 time steps.}
  \label{vis}
  \vskip -0.1 in
\end{figure}

Figure \ref{vis} shows some visualization results of object detection using our spiking neural network, with images from both the MS COCO and our spike dataset. In this experiment, we set all time steps to 300. Specifically, for the MS COCO, when performing detection, we will convert the images to spikes first, then use SNN to perform detection and output the detection results to the original images. For our spike dataset, we directly perform detection on the spike data and visualize the results to their corresponding images.
\vskip -0.1 in
\section{Conclusion}
In this paper, we present a low latency SNN generation method by object detection task for accurate conversion with low latency. We theoretically analyze the error of the ANN-SNN conversion process and illustrate the influence of the quantization error and spike firing rate on the accurate conversion. Then we propose low spike firing rate layer replacement method. It significantly reduces the problem of firing rate due to spike feature scale variation in SNN networks, thus reducing the inference time step required for the network. To address the problem of quantization error, we propose the mechanism of quantization activation function Quant-ReLU and residual fix to alleviate the above problem. The experimental results show that the time step required for SNN network inference is greatly reduced after using the above method, thus greatly reducing the real-time delay of the detection network. Beyond the object detection task, the proposed methods are theoretically generalizable to other SNN tasks.

\bibliographystyle{IEEEbib}
\bibliography{main}

\end{document}